\pdfoutput=1

\documentclass[11pt]{article}

\usepackage[final]{coling}

\usepackage{times}
\usepackage{latexsym}
\usepackage{graphicx}
\usepackage{multirow}
\usepackage{amssymb}
\usepackage{bbding}
\usepackage{amsmath}
\usepackage{mathrsfs}
\usepackage{hyperref}
\usepackage{caption,subcaption}

\usepackage[T1]{fontenc}

\usepackage[utf8]{inputenc}

\usepackage{microtype}

\usepackage{inconsolata}

\usepackage{graphicx}

\title{MoSLD: An Extremely Parameter-Efficient Mixture-of-Shared LoRAs for Multi-Task Learning}

\author{Lulu Zhao$^{1}$\thanks{\ \ \ Lulu Zhao is the corresponding author.}, Weihao Zeng$^{2}$, Xiaofeng Shi$^{1}$, Hua Zhou$^{1}$ \\
 $^1$Beijing Academy of Artificial Intelligence (BAAI) \\
  $^{2}$School of Artificial Intelligence, Beijing University of Posts and Telecommunications\\
  \texttt{\{llzhao\}@baai.ac.cn}\\
  }

\begin{document}
\maketitle
\begin{abstract}
Recently, LoRA has emerged as a crucial technique for fine-tuning large pre-trained models, yet its performance in multi-task learning scenarios often falls short. In contrast, the MoE architecture presents a natural solution to this issue. However, it introduces challenges such as mutual interference of data across multiple domains and knowledge forgetting of various tasks. Additionally, MoE significantly increases the number of parameters, posing a computational cost challenge. Therefore, in this paper, we propose MoSLD, a mixture-of-shared-LoRAs model with a dropout strategy. MoSLD addresses these challenges by sharing the upper projection matrix in LoRA among different experts, encouraging the model to learn general knowledge across tasks, while still allowing the lower projection matrix to focus on the unique features of each task. The application of dropout alleviates the imbalanced update of parameter matrix and mitigates parameter overfitting in LoRA. Extensive experiments demonstrate that our model exhibits excellent performance in both single-task and multi-task scenarios, with robust out-of-domain generalization capabilities.
\end{abstract}

\section{Introduction}

\begin{figure}[t]
\centering
\resizebox{0.45\textwidth}{!}{
\includegraphics[scale=0.5]{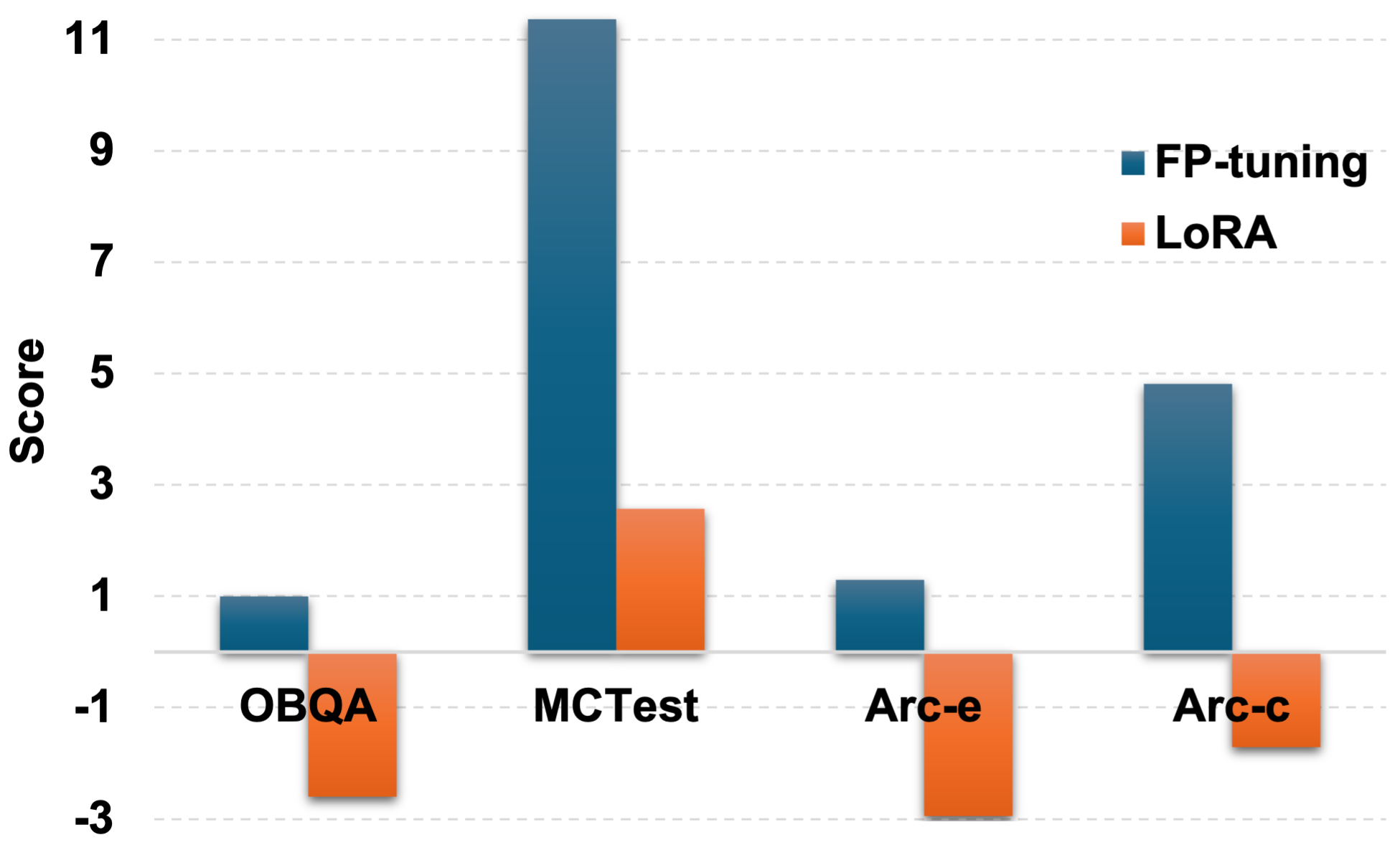}}
\caption{The increase between mixture setting and single setting for FP-tuning and LoRA on four datasets. The vertical axis is Score (mixture)-Score (single).}
\label{fig:intro}
\vspace{-0.5cm}
\end{figure}

The emergence of Large Language Models (LLMs) has significantly advanced Natural Language Processing (NLP) technology, serving as a robust foundation with broad applicability \cite{touvron2023llama1, touvron2023llama, NEURIPS2022_b1efde53}. However, as the parameter scale increases, the process of full parameter fine-tuning (FP-tuning) demands substantial computational and memory resources. To strike a balance between resource requirements and effectiveness, the research community is increasingly turning to parameter-efficient fine-tuning (PEFT) methods \cite{10.1145/3477495.3531933, zeng-etal-2023-seen}, with LoRA emerging as the most prevalent and effective choice. Nevertheless, training an LLM via LoRA with multi-faceted capabilities faces significant challenges due to the differences and diversity inherent in various tasks. Figure \ref{fig:intro} illustrates that while FP-tuning demonstrates competitive performance in a multi-task mixed training data setting, plain LoRA exhibits a drop. This decline underscores the challenge posed by the heterogeneity and imbalance in training data, resulting in interference between data from different tasks and consequently degrading the performance of plain LoRA on in-domain tasks. In essence, plain LoRA proves highly sensitive to the configuration of training data.

As we all know, MoE \cite{shazeer2017} has demonstrated remarkable advantages in amalgamating multiple capabilities. Particularly, the integration of MoE and LoRA \cite{hu2022lora} stands out as a promising approach to leveraging MoE in a parameter-efficient manner. This method preserves domain knowledge while significantly reducing training costs by introducing a limited number of domain-specific parameters \cite{dou2024loramoe,luo2024moelora,liu2023moelora}. Presently, several works are devoted to applying MoE to LoRA. Some directly combine trained LoRAs linearly \cite{zhang2023composing, huang2024lorahub}, while others apply combinations of MoE and LoRA to different backbones \cite{chen2024llavamole, dou2024loramoe}. Another approach involves training a LoRA module for each distinct task type and employing a routing mechanism to integrate the LoRA modules under a shared LLM \cite{feng2024mixtureofloras}. However, we contend that these methods inadequately address the issue of data conflicts across different domains during LoRA training. Three primary challenges emerge: (1) The MoE architecture emphasizes the unique attributes of each LoRA and overlooks the transfer of general knowledge between different LoRAs, thereby impeding cross-task generalization in LLMs; (2) Requires a large number of trained LoRA modules (for each task); (3) Multiple LoRAs escalate the number of parameters and computational costs.

To solve these issues, in this paper, we propose a parameter-sharing method applied to the mixture-of-LoRAs, called MoSLD. The plain LoRA module comprises the upper projection matrix (A) and the lower projection matrix (B), which can be viewed as naturally decoupled general-feature and specific-feature matrices, respectively. Building upon the classic MoE architecture, we enable all experts at each layer to share a general-feature matrix while retaining the specific-feature matrix of each expert. This approach compels the model to capture shared general knowledge across various tasks to the fullest extent. The shared operation notably reduces the parameters of the MoE architecture, aligning with findings indicating parameter redundancy among experts \cite{Fedus, kim2021scalable}. Despite the majority of parameters in the LoRA module being shared, differences can still be learned in each expert's specific-feature matrix due to the tight coupling between the general and specific features. We posit that this mechanism can adaptively generalize to any new task. Furthermore, recognizing that the general-feature matrix is updated more frequently than the specific-feature matrix during fine-tuning, and overfitting tends to occur in LoRA \cite{wang2024lora}, we apply the dropout strategy to the general-feature matrix, that is some weight values are randomly set to zero during training. This approach helps balance the updates between the general-feature and specific-feature matrices. Consequently, it not only facilitates a more balanced information exchange between different experts but also mitigates issues related to parameter redundancy and optimization imbalance.

In summary, our contributions are as follows: (1) We introduce a parameter-efficient MoSLD approach that disentangles domain knowledge and captures general knowledge by sharing a general-feature matrix, thus mitigating interference between heterogeneous datasets. (2) We implement a dropout strategy on the general-feature matrix to effectively mitigate overfitting and address the imbalance in directly optimizing MoE. (3) We conduct extensive experiments on various benchmarks to validate the effectiveness of our methods. Additionally, our approach demonstrates superior generalization to out-of-domain data.

\begin{figure*}[t]
\centering
\includegraphics[width=15cm, height=5cm]{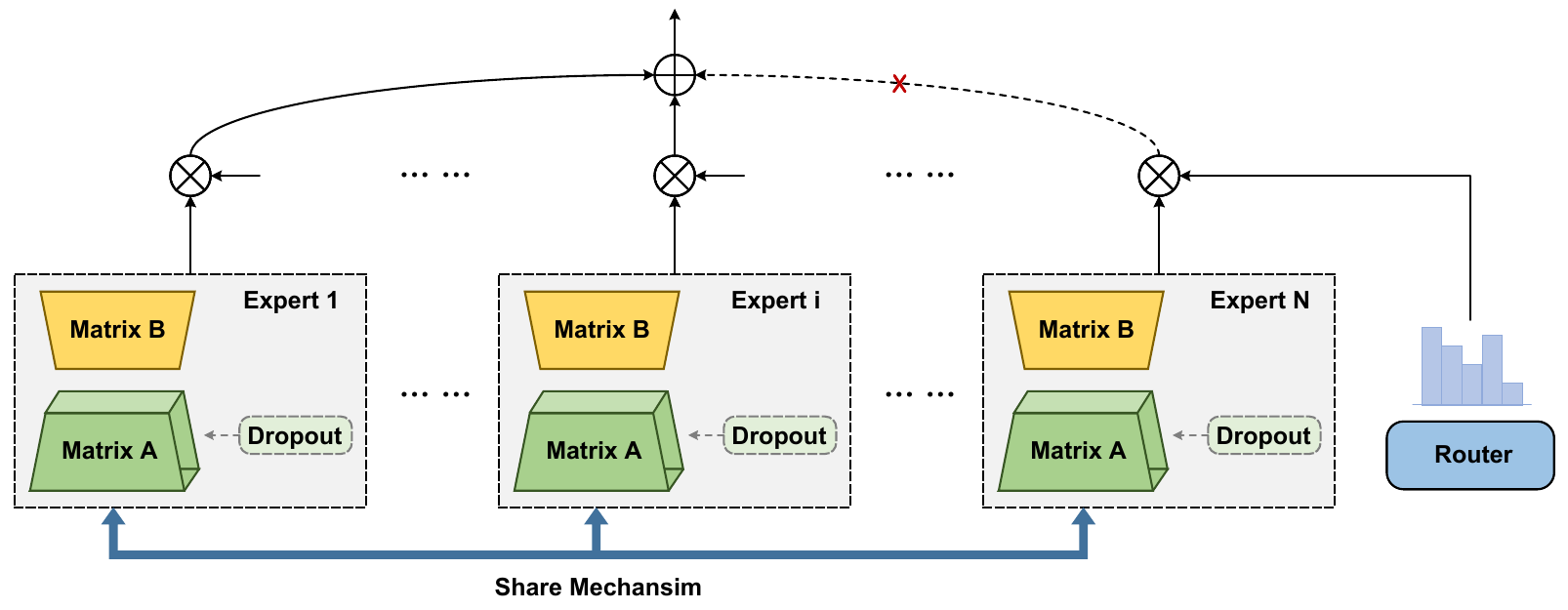}
\caption{Overview of the share mechansim and dropout strategy in our MoSLD. Noted that the matrix A is shared among all experts in each layer.}
\label{fig:share}
\vspace{-0.5cm}
\end{figure*}

\section{Related Work}
\subsection{Mixture-of-Expert}
The Mixture of Experts (MoE) functions as an ensemble method, conceptualized as a collection of sub-modules or experts, each tailored to process distinct types of input data. Guided by a router, each expert is selectively activated based on the input data type. This technique has garnered increasing attention and demonstrated remarkable performance across various domains, including computer vision, speech recognition, and multi-modal applications \cite{fedus2022review}. Evolution of MoE techniques spans from early sample-level approaches \cite{6797059} to contemporary token-level implementations \cite{shazeer2017,NEURIPS2021_48237d9f}, which have now become mainstream. Concurrently, some researchers \cite{NEURIPS2022_2f00ecd7,chi2022on} are delving into the router selection problem within MoE. Notably, the majority of these endeavors aim to scale up model parameters while mitigating computational costs.

\subsection{Mixture-of-LoRA}
As LoRA gradually becomes the most common parameter-efficient fine-tuning method, researchers pay more attention to combining MoE and LoRA for more efficient and effective model tuning. \citet{huang2024lorahub} and \citet{feng2024mixtureofloras} pioneer the approach of training several LoRA weights on upstream tasks and then integrating the LoRA modules into a shared LLM using a routing mechanism. However, these methods necessitate the training of numerous pre-defined LoRA modules. \citet{chen2024llavamole} initially engage in instruction fine-tuning through sparse mixing of LoRA experts in the multi-modal domain, while \citet{dou2024loramoe} split the LoRA experts into two groups to explicitly learn different capabilities for each group. These mixture-of-LoRA methods typically involve predefined hyperparameters that require careful selection, and they densely mix LoRA experts, significantly increasing computational costs. To tackle overfitting resulting from an excessive number of experts, \citet{gao2024higher} allocate a varying number of experts to each layer. \citet{wu2024mixture} propose MOLE, treating each layer of trained LoRAs as a distinct expert and implementing hierarchical weight control through a learnable gating function within each layer to tailor composition weights specific to a given domain's objectives. However, these approaches overlook the issue of data conflicts across different datasets during LoRA training. As our concurrent work, MixLoRA \cite{li2024mixloraenhancinglargelanguage} also focuses on multi-task learning, which fuses multiple LoRAs with the shared FFN layer and employ a plain LoRA on the self-attention layer. We believe this method will introduce a large number of additional trainable parameters and incur a huge computational cost. In our study, we conduct extensive experimental analysis for both single and mixture data settings in a more lightweight way.

\section{Methodology}
In this section, we describe our MoSLD from the sharing mechanism, dropout strategy and optimization details, as shown in Figure \ref{fig:share}.

\subsection{Sharing Mechanism of LoRAs}
In the area of parameter-efficient fine-tuning, LoRA introduces the concept of training only two low-rank matrices as an alternative to dense layer updates. In other words, it reformulates the parameter fine-tuning process in LLMs as a low-rank decomposition. Specifically, the equation $W_0+\Delta{W} = W_0+BA$ captures this decomposition. Here, $W_0 \in \mathcal{R}^{d_{in} \times d_{out}}$  represents the parameter matrix of the pre-trained LLM, while $\Delta{W} \in \mathcal{R}^{d_{in} \times d_{out}}$ denotes the matrix updated during fine-tuning. The matrices $B \in \mathcal{R}^{d_{in} \times r}$  and $A \in \mathcal{R}^{r \times d_{out}}$ are low-rank and trainable.

In order to achieve the transfer of general features between different tasks and capture the shared general knowledge, we design a novel sharing mechanism. Specifically, given a Transformer model with $L$ layers, we allocate $N_l$ experts for layer $l$ and create $N_l$ pairs of low-rank matrices $\{A_{i, l}, B_{i, l}\}_{i=1}^{N_l}$, where $A_{i, l}$ is initialized from a random Gaussian distribution and each $B_{i, l}$ is set to zero. It is worth noting that the matrix $A_{i, l}$ is shared among all experts in each layer, i.e., $A_{1, l}=A_{2, l}...=A_{N_l, l}$ \ $(l \in L)$. In other words, the core idea is to share the matrix A as the general-feature matrix and keep matrix B as specific-feature matrix. In this way, we can only keep $L$ central general-feature matrices for a L-layer MoE architecture, which significantly reduces the parameters of the MoE architecture. A router with a trainable weight matrix $W_l \in \mathcal{R}^{d_{in} \times N_l}$ is used to specify different experts for the input $x$. As in the original MoE, MoSLD selects the top $K$ experts for computation, and the gate score $S_l^k$ is calculated as follows:

\begin{equation}
\begin{aligned}
S_l^k(x) = \frac{\operatorname{TopK}(\operatorname{softmax}(W_lx),K)_k}{\sum_{k=1}^{K}\operatorname{TopK}(\operatorname{softmax}(W_lx),K)_k}
\end{aligned}
\end{equation}

\subsection{Dropout Strategy}
In order to alleviate the imbalance and over-fitting problems caused by frequent general-feature matrix updates, we propose to apply the dropout strategy on the general-feature parameter matrix $A_l$. Dropout involves randomly ignoring a proportion of updates to the parameter matrix during each iteration of training. This technique helps prevent over-reliance on specific parameters and encourages robust learning by introducing noise. That is, at each iteration, we take a certain probability $p$ to discard the update in the general-feature matrix. Specifically, we generate a binary mask matrix drawn from Bernoulli distribution with a mask probability $p$, where each element in the general-feature matrix independently takes a value of 1 (keeping the parameter) with probability $1-p$ or 0 (dropping the parameter) with probability $p$.  The general-feature matrix is updated as follows:
\begin{equation}
\begin{aligned}
\operatorname{Mask} \sim \operatorname{Bernoulli}(p) \\
\mathbf{A}^{'}_{l} = \operatorname{Mask} \odot \mathbf{A}_{l} \\
\widetilde{\mathbf{A}^{'}}_{l} = \mathbf{A}^{'}_{l}/(1-p)
\end{aligned}
\end{equation}

\begin{figure}[t]
\centering
\includegraphics[width=6cm, height=6cm]{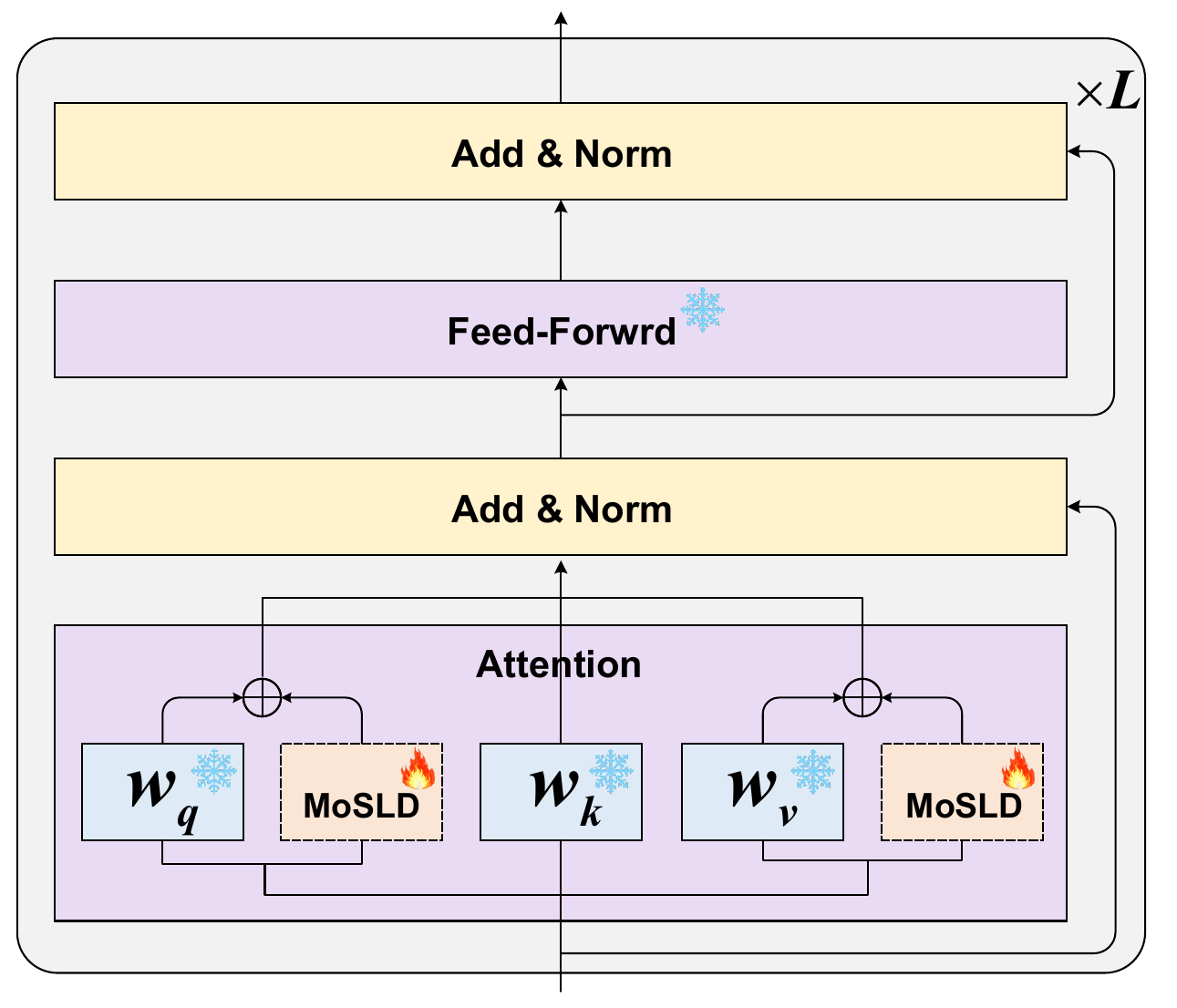}
\caption{The overview of our proposed Mixture-of-Shared-LoRA with dropout strategy applied on $W_q$ and $W_v$.
}
\label{fig:moe}
\vspace{-0.5cm}
\end{figure} 

\subsection{The Overall Procedure}
Our method is a combination of shared LoRA modules and MoE framework, as shown in Figure \ref{fig:moe}. Here, we apply our MoSLD on the matrix $Q$ and matrix $V$ of the self-attention layer:

\begin{equation}
\begin{aligned}
h_l = W_0x+\frac{\alpha}{r}\sum_{k=1}^{K}S_l^k(x)A_{k,l}B_{k,l}x
\end{aligned}
\end{equation}
where $W_0 \in \{W_q, W_v\}$ and $h_l$ is the output embedding. Besides, similar to previous sparse MoE works, the load balancing loss $L_b$ is also applied on each MoE layer, which is formulated as:

\begin{equation}
\begin{aligned}
L_{b} = \sum_{k=1}^{K}c_k\cdot s^{k} \\
p_k = \sum_{x\in X}\frac{e^{S^k(x)}}{\sum^{k}e^{S^k(x)}}
\end{aligned}
\end{equation}
where $c_k$ is the number of tokens assigned to the $k$-th expert.

\begin{table*}[]
\centering
\resizebox{1.0\textwidth}{!}{%
\begin{tabular}{l|c|cccccc|c}
\hline
\multicolumn{2}{l|}{\textbf{Model}} & \textbf{OBQA} & \textbf{CSQA} & \textbf{Race} & \textbf{MCTest} & \textbf{Arc-e} & \textbf{Arc-c} & \textbf{Avg} \\
\hline
\multirow{2}{*}{\textbf{FP-tuning}} & single & 75.00 &75.74 & 80.62 & 39.05 & 72.39 & 60.63 & 67.24\\
& mixture & 76.00 & 75.27 & 81.46 & 50.42 & 73.69 & 65.45 & 70.38\\
\hline
\multirow{2}{*}{\textbf{Prefix-tuning}} & single & 47.76 &42.65 & 53.77 & 25.19 & 45.65 & 35.50 & 41.70\\
& mixture & 46.51 & 44.98 & 49.88 & 22.46 & 47.92 & 35.30 & 41.18 \\
\hline
\multirow{2}{*}{\textbf{LoRA}} & single & 75.40 & 76.33 & 76.06 & 53.10 & 73.82 & 62.71 & 69.57\\
& mixture & 72.80 & 76.30 & 78.23 & 55.67 & 70.87 & 61.00 & 69.15 \\
\hline
\multirow{2}{*}{\textbf{MoLoRA}} & single & 74.71 & 76.65 & 74.26 & 49.08 & 74.14 & 61.38 & 68.37\\
& mixture & 75.04 & 75.27 & 73.88 & 55.37 & 75.25 & 62.86 & 69.61 \\
\hline
\multirow{2}{*}{\textbf{SiRA}} & single & 73.99  &  76.26 & 75.63 & 48.28 & 74.02 & 62.86 & 68.51\\
& mixture & 74.34 & 76.22 & 75.04 & 52.33 & 74.98 & 63.16 & 69.35 \\
\hline
\multirow{2}{*}{\textbf{MoLA}} & single & 74.60 & 77.23 & 75.29 & 44.90 & 72.73 & 60.80 & 67.59\\
& mixture & 76.60 & 73.46 & 75.25 & 54.42 & 76.34 & 63.91 & 70.00 \\
\hline
\multirow{2}{*}{\textbf{MixLoRA}} & single & 75.60 & 74.83 & 75.47 & 50.88 & 74.51 & 60.10 & 68.57\\
& mixture & 75.80  & 76.81 & 74.79 & 54.26 & 74.41 & 63.62 & 69.95 \\
\hline
\multirow{2}{*}{\textbf{MoSL (our)}} & single & 76.30 & 77.56 & 74.63 & 49.66 & 76.30 & 60.48 & 69.16\\
& mixture & \textbf{76.80} (+0.50) & 75.02 (-2.54) & \textbf{74.69} (+0.06) & \textbf{58.50} (+8.84) & 76.09 (-0.21) & \textbf{64.16} (+3.68) & \textbf{70.88} (+1.72) \\
\hline
\multirow{2}{*}{\textbf{MoSLD (our)}} & single & 78.40 & 75.84 & 76.08 & 53.06 & 76.35 & 61.49 & 70.20\\
& mixture & \textbf{78.80} (+0.40) & \textbf{76.43} (+0.59) & \textbf{76.96} (+0.88)  & \textbf{54.42} (+1.36) & \textbf{76.60} (+0.25)  & \textbf{66.13} (+4.64)  & \textbf{71.56} (+1.36) \\
\hline
\end{tabular}%
}
\caption{Results of different methods on the in-domain test sets of six commonsense reasoning datasets. We also report the increase of mixture setting compared to single setting. Results are averaged over three random runs. (p < 0.01 under t-test)}
\label{Table: main_results}
\vspace{-0.5cm}
\end{table*}

\section{Experimental Setup}
\subsection{Datasets}

To evaluate the effectiveness of MoSLD, we conduct experiments on six commonsense reasoning datasets, including commonsense QA task (OBQA \cite{OpenBookQA2018}, CSQA \cite{talmor-etal-2019-commonsenseqa}), reading comprehension task (Race \cite{lai-etal-2017-race}, MCTest \cite{richardson-etal-2013-mctest}), and subject knowledge QA task (Arc-e \cite{allenai:arc}, and Arc-c \cite{allenai:arc}). We denote the six datasets as $\{D_1, \ D_2, ..., \ D_6\}$, and we also  create a mixed dataset $D_{mix}$, corresponding to the single setting and the mixture setting respectively. The dataset sizes are as follows for training and testing: 5,457/500, 10,962/1140, 10,083/4934, 1,330/147, 2,821/2,376, and 1,418/1,172. We allocate 10\% of the training set for validation. For all datasets, we use answer accuracy as the evaluation metric.

\subsection{Baselines}
We compare MoSLD with five parameter-efficient fine-tuning methods: Prefix-tuning \cite{li-liang-2021-prefix, zhao-etal-2022-domain}, LoRA \cite{hu2022lora}, MoLoRA \cite{zadouri2024pushing}, SiRA \cite{zhu2023sirasparsemixturelow}, MoLA \cite{gao2024higher}, MixLoRA \cite{li2024mixloraenhancinglargelanguage}. Additionally, we evaluate full-parameter fine-tuning. The details can be seen in Appendix \ref{sec:baseline}.






\subsection{Training Details}

We take LLaMA2-7B \cite{touvron2023llama} which contains 32 layers as our base model. For plain LoRA and its variants, the $r$ is set to 8 and $\alpha$ is 16. Beside, the LoRA modules are used in matrix Q and matrix V in attention layers. Our MoSLD also follows the same settings. We allocate 8 experts to each layer for 1-8 layers, 6 experts to each layer for 9-16 layers, 4 experts to each layer for 17-24 layers, and 2 experts to each layer for the last 8 layers. The K of the selected experts is 2. For training details, we finetune models with 10 epochs and a peak of 3e-4 learning rate. The drop ratio applied to matrix A is set to 0.1. The batch size during model tuning is 128.  The experiments are run on 16 NVIDIA A100 40GB GPUs. 

\begin{figure*}[t]
\centering
\subcaptionbox{OBQA\label{openbookqa}}{\includegraphics[width =0.28\linewidth]{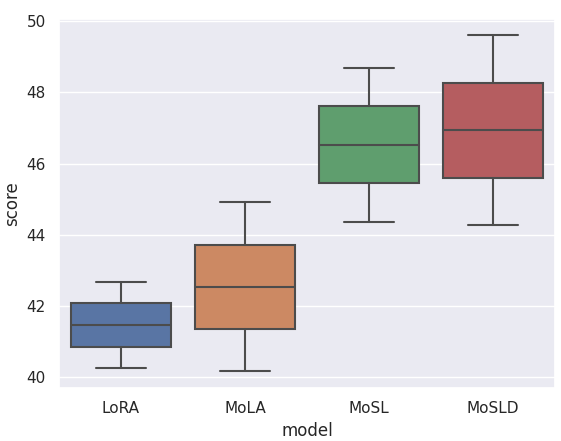}}\hfill
\subcaptionbox{CSQA\label{commonsenseqa}}{\includegraphics[width =0.28\linewidth]{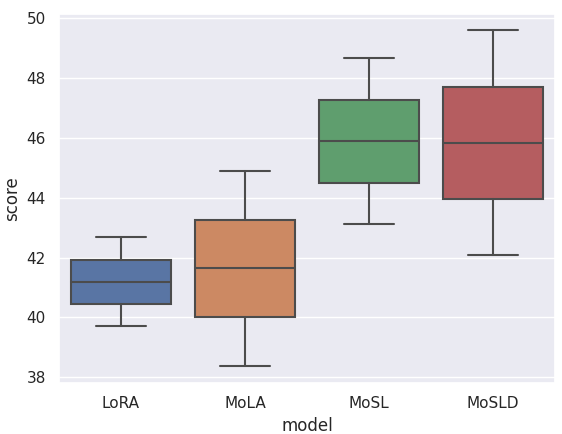}}\hfill
\subcaptionbox{Race\label{Race}}{\includegraphics[width =0.28\linewidth]{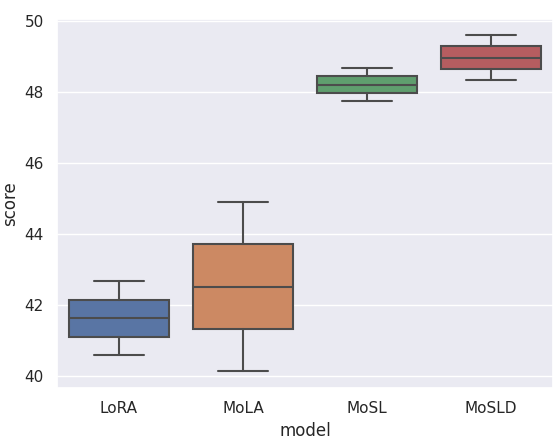}}

\subcaptionbox{MCTest\label{McTest}}{\includegraphics[width =0.28\linewidth]{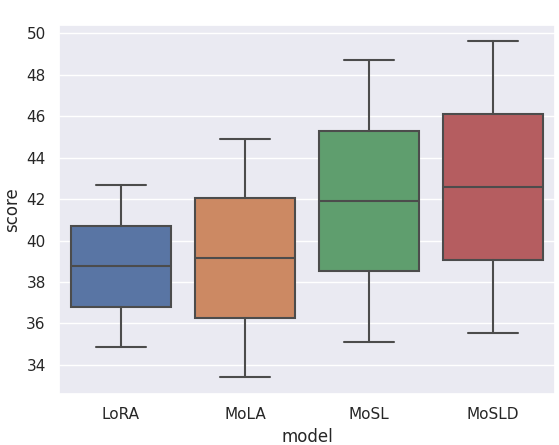}}\hfill
\subcaptionbox{Arc-e\label{Arc-e}}{\includegraphics[width =0.28\linewidth]{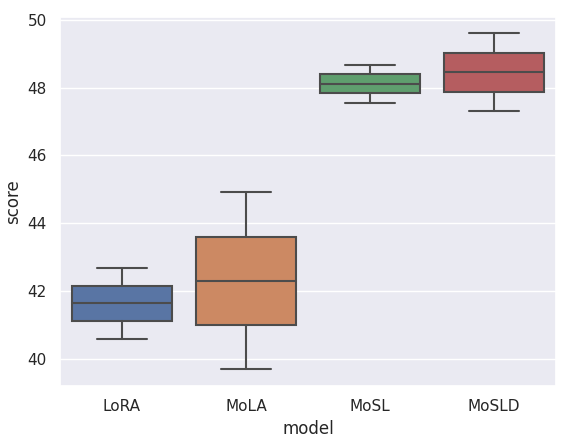}}\hfill
\subcaptionbox{Arc-c\label{Arc-c}}{\includegraphics[width =0.28\linewidth]{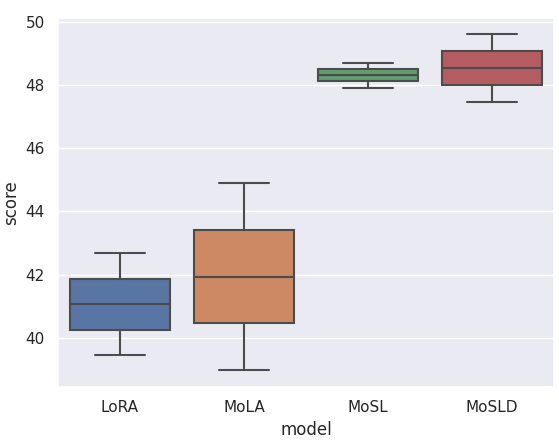}}
\caption{A comparision of performance for LoRA, MoLA, MoSL, and MoSLD on single and mixture settings for MMLU test set.}
\label{Figure: single_mixture}
\vspace{-0.3cm}
\end{figure*}

\subsection{Main Results}
Table \ref{Table: main_results} presents the experimental outcomes of various baselines under both single and mixture settings across different datasets. Initially, we report the performance of models trained on individual datasets. LoRA notably outperforms other baselines, exhibiting improvements of 2.33\% and 27.87\% over FP-tuning (single) and Prefix-tuning (single), respectively. MoLoRA, SiRA, MoLA, and MixLoRA trail behind LoRA by 1.20\%, 1.06\%, 1.98\%, and 1.00\%, indicating that simply combining LoRA and MoE does not confer an advantage in single in-domain datasets. After establishing a robust baseline in the single setting, we proceed to report results for the mixture setting. Here, we observe a decline in LoRA's performance, trailing 1.23 points behind FP-tuning (70.38\%). Conversely, applying the MoE framework to LoRA, i.e., MoLoRA, SiRA, MoLA, and MixLoRA, achieves scores of 69.61\%, 69.35\%, 70.00\%, and 69.95\%, demonstrating MoE's suitability for multi-task scenarios and MoLA is the best performing baseline in the mixture setting. Further comparison between single and mixture settings reveals that FP-tuning and MoLA improve by 3.14\% and 2.41\%, respectively, in the mixture setting compared to the single setting. However, LoRA's performance decreases by 0.42\% in the mixture setting compared to the single setting, indicating conflicts between multi-task data and the mixture strategy's detrimental impact on performance.

Upon closer examination, our proposed MoSLD demonstrates performance enhancements of 2.61\% and 1.56\% over MoLA in single and mixture settings, respectively. This emphasizes the effectiveness of the sharing mechanism and dropout strategy in alleviating data conflicts and retaining shared knowledge between various tasks. Furthermore, conducting ablation experiments by removing the dropout strategy, MoSL experiences performance decreases of 1.04\% and 0.68\%, respectively, compared to MoSLD. This highlights the crucial role of the dropout strategy in mitigating training overfitting and optimization imbalance. Nevertheless, MoSL still achieves competitive results of 69.16\% and 70.88\%. We also found that our model not only achieves good results in the mixture setting, but also achieves good results in the single setting, which overcomes the disadvantage of MoLA's poor performance in the single setting. However, we find that our models, especially MoSL, do not have much advantage over plain LoRA, which is consistent with the performance of all baselines combining MoE with LoRA. This is because the complexity of the model ensemble causes overfitting on a single simple task, resulting in little improvement. In conclusion, our approach exhibits significant advantages under both single and mixture settings, particularly in alleviating data conflicts across multiple tasks and addressing knowledge forgetting issues in multi-task learning. In addition, we also pay attention to the efficiency of training. Due to the introduction of multiple LoRAs, the trainable parameters of MoLA are higher than those of plain LoRA. However, although our MoSLD expands LoRA several times through the MoE architecture, it does not introduce a large number of additional parameters and also enables the LoRA training to have multiple capabilities. Details can be seen in Section \ref{sec:cost}.

\begin{table*}[htp]
\centering
\resizebox{0.8\textwidth}{!}{%
\begin{tabular}{l|c|cccccc|c}
\hline
\multicolumn{2}{l|}{\textbf{Model}} & \textbf{OBQA} & \textbf{CSQA} & \textbf{Race} & \textbf{MCTest} & \textbf{Arc-e} & \textbf{Arc-c} & \textbf{Avg} \\
\hline
\multirow{2}{*}{\textbf{LoRA}} & single & 75.40 & 76.33 & 76.06 & 53.10 & 73.82 & 62.71 & 69.57\\
& mixture & 72.80 & 76.30 & 78.23 & 55.67 & 70.87 & 61.00 & 69.15 \\
\hline
\multirow{2}{*}{\textbf{MoLA}} & single & 74.60 & 77.23 & 75.29 & 44.90 & 72.73 & 60.80 & 67.59\\
& mixture & 76.60 & 73.46 & 75.25 & 54.42 & 76.34 & 63.91 & 70.00 \\
\hline
\multirow{2}{*}{\textbf{MoSLD (matrix A)}} & single & 78.40 & 75.84 & 76.08 & 53.06 & 76.35 & 61.49 & \textbf{70.20}\\
& mixture & 78.80 & 76.43 & 76.96 & 54.42 & 76.60 & 66.13 & \textbf{71.56} \\
\hline
\multirow{2}{*}{\textbf{MoSLD (matrix B)}} & single & 77.60 & 75.76 & 74.58 & 46.94 & 76.09 & 60.83 & 68.63\\
& mixture & 76.40 & 74.11 & 75.25 & 56.46 & 77.15 & 65.02 & 70.73 \\
\hline
\end{tabular}%
}
\caption{The results for applying our methods on matrix A and matrix B.}
\label{Table: Location}
\vspace{-0.3cm}
\end{table*}

\section{Qualitative Analysis}

\subsection{Out-of-domain Test}
To assess the generalization capability of our proposed model, we conducted out-of-domain experiments using the test set of MMLU. Figure \ref{Figure: single_mixture} presents a boxplot, where the top and bottom horizontal lines represent the mixture and single settings, respectively. Our models, MoSL and MoSLD, consistently outperform others in both settings, exhibiting significant improvements, particularly on Race, Arc-e, and Arc-c datasets. This highlights the effectiveness of our models in disentangling domain knowledge and transferring general features across diverse datasets. OBQA and CSQA exhibit similar trends in the boxplot, indicating similar data distributions between the two datasets. Conversely, for MCTest, while improvements are observed in the mixture settings, the single settings remain relatively unchanged. This divergence may stem from the substantial differences between the MCTest and MMLU test sets, suggesting that introducing data from other domains or tasks could inspire general domain knowledge. In summary, our model demonstrates strong generalization capabilities, particularly in multi-task scenarios.

\begin{figure}[htbp]
\centering
\subcaptionbox{OBQA\&CSQA\&Race\label{three_f}}{\includegraphics[width =0.48\linewidth]{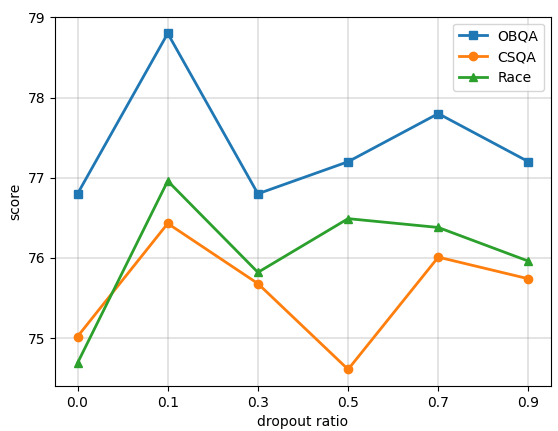}}\hfill
\subcaptionbox{MCTest\&Arc-e\&Arc-c\label{three_l}}{\includegraphics[width =0.48\linewidth]{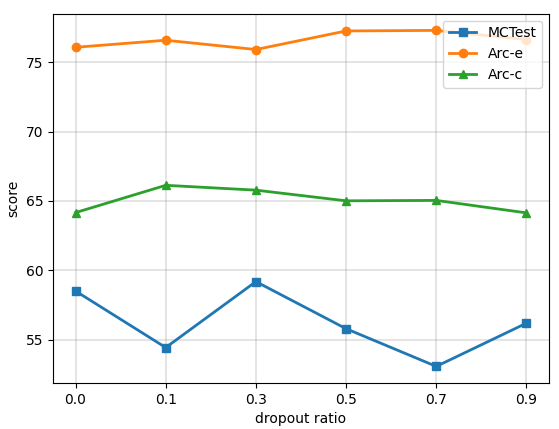}}
\caption{Results of six datasets under different dropout ratios. Here, we are based on the mixture setting.}
\label{Figure: dropout}
\vspace{-0.5cm}
\end{figure}

\begin{figure*}[t]
\centering
\subcaptionbox{OBQA\label{openbookqa}}{\includegraphics[width =0.28\linewidth]{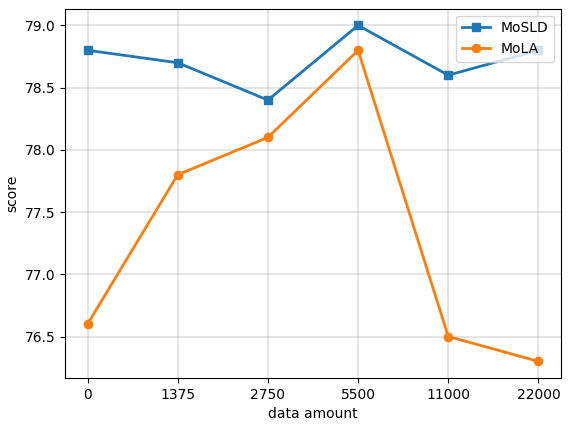}}\hfill
\subcaptionbox{CSQA\label{commonsenseqa}}{\includegraphics[width =0.28\linewidth]{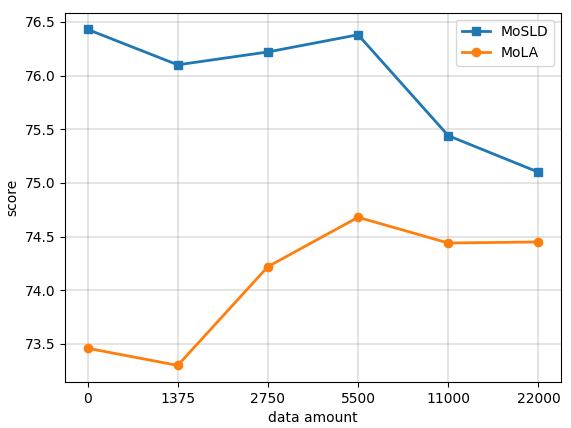}}\hfill
\subcaptionbox{Race\label{Race}}{\includegraphics[width =0.28\linewidth]{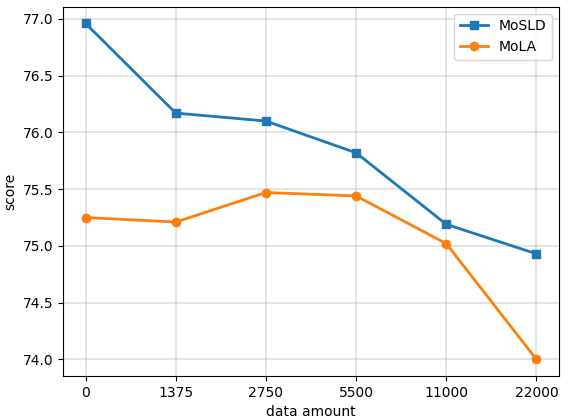}}

\subcaptionbox{MCTest\label{McTest}}{\includegraphics[width =0.28\linewidth]{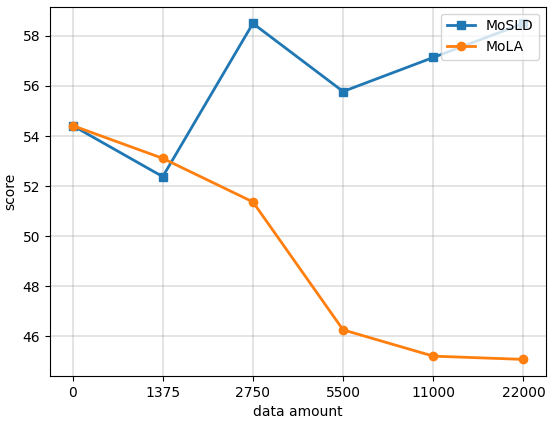}}\hfill
\subcaptionbox{Arc-e\label{Arc-e}}{\includegraphics[width =0.28\linewidth]{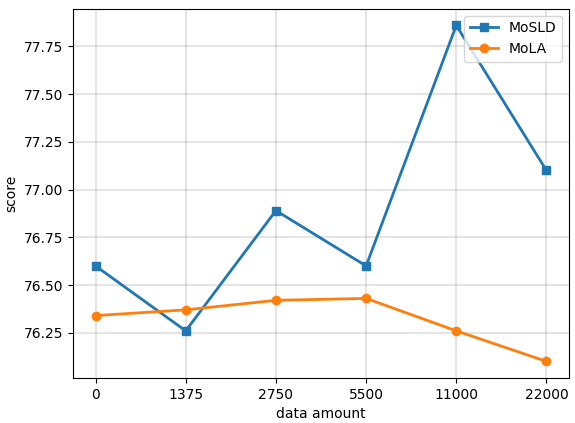}}\hfill
\subcaptionbox{Arc-c\label{Arc-c}}{\includegraphics[width =0.28\linewidth]{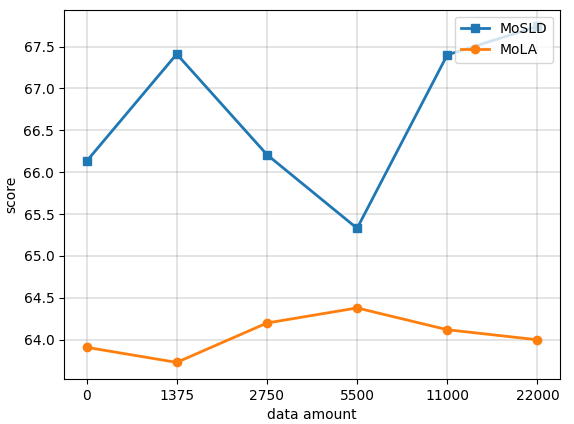}}
\caption{Different data amount of OpenOrca between MoSLD and MoLA on six datasets. Here, we use the mixture setting.}
\label{Figure: data_amount}
\vspace{-0.3cm}
\end{figure*}

\begin{figure}[htbp]
\centering
\resizebox{0.45\textwidth}{!}{
\includegraphics[scale=0.5]{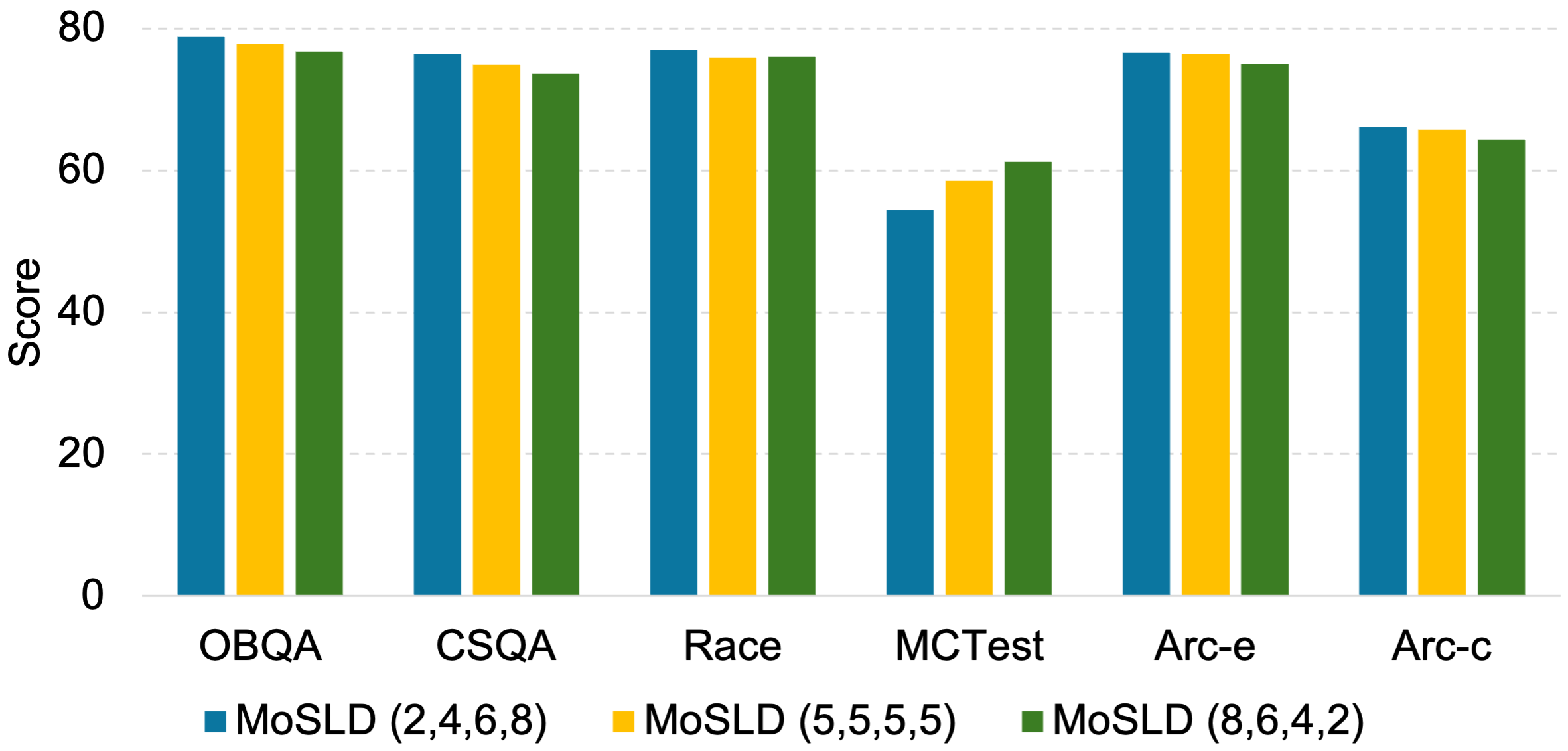}}

\caption{Different allocation strategies for the number of experts at different layers of the model. Here, we use the mixture setting.}
\label{fig:expert}
\vspace{-0.7cm}
\end{figure} 

\subsection{Effect of Model Parameters}
In this section, we conduct parameter search experiments. 

\noindent\textbf{Dropout Location} \quad As shown in Table \ref{Table: Location}, we show the results of applying our methods on matrix A and matrix B. We found that in the single setting, MoSLD (matrix B) does not achieve much improvement, 0.94 points lower than the ordinary LoRA and 1.04 points higher than MoLA. The mixture setting still achieves good results. However, the results of applying our method on matrix B are lower than those of applying it on matrix A in both the single and mixture settings. This also shows that matrix A is more used to extract general features.

\noindent\textbf{Dropout Ratio} \quad In Figure \ref{Figure: dropout}, we depict the performance of six datasets under the mixture setting with varying dropout ratios. We observe a general downward trend in most results as the dropout ratio increases. This phenomenon occurs because while dropout can mitigate overfitting to some extent, excessively high dropout rates may diminish the model's capabilities. Therefore, careful selection of the dropout ratio parameter is necessary. Interestingly, the curves for the Arc-e and Arc-c datasets remain relatively stable across different dropout ratios. We attribute this stability to the simplicity of these two datasets, where model sparsification has minimal impact on the results.

\begin{table*}[]
\centering
\resizebox{0.7\textwidth}{!}{%
\begin{tabular}{l|c|cccc|c}
\hline
\multicolumn{2}{l|}{\textbf{Model}} & \textbf{OBQA} & \textbf{MCTest} & \textbf{Arc-c} & \textbf{GSM8K} & \textbf{Avg} \\
\hline
\multirow{2}{*}{\textbf{LoRA}} & single & 75.40 & 53.10 & 62.71 & 23.12 & 53.58\\
& mixture & 73.20 & 55.10 & 64.08 & 17.51 & 52.47 \\
\hline
\multirow{2}{*}{\textbf{MoSLD}} & single & 78.40 & 53.06 & 61.49 & 22.06 & 53.75\\
& mixture & 79.80 & 53.90 & 63.29 & 22.73 & 53.93 \\
\hline
\end{tabular}%
}
\caption{The results of the mixture setting of tasks with little commonality.}
\label{Table: little_commonality}
\vspace{-0.3cm}
\end{table*}

\begin{table*}[]
\centering
\resizebox{0.8\textwidth}{!}{%
\begin{tabular}{l|c|cccccc|c}
\hline
\multicolumn{2}{l|}{\textbf{Model}} & \textbf{OBQA} & \textbf{CSQA} & \textbf{Race} & \textbf{MCTest} & \textbf{Arc-e} & \textbf{Arc-c} & \textbf{Avg} \\
\hline
\multirow{2}{*}{\textbf{LLaMA2-7B}} & single & 78.40 & 75.84 & 76.08 & 53.06 & 76.35 & 61.49 & 70.20\\
& mixture & 78.80 & 76.43 & 76.96 & 54.42 & 76.60 & 66.13 & 71.56\\
\hline
\multirow{2}{*}{\textbf{LLaMA2-13B}} & single & 81.4 & 77.95 & 78.01 & 57.86 & 78.93 & 65.05 & 73.20\\
& mixture & 82.2 & 78.46 & 79.87 & 58.50 & 79.67 & 70.14 & 74.81 \\
\hline
\multirow{2}{*}{\textbf{LLaMA-33B}} & single & 83.93 & 81.49 & 83.27 & 65.99 & 85.10 & 68.52 & 78.05\\
& mixture & 84.55 & 83.26 & 84.90 & 66.73 & 85.95 & 74.36 & 79.96 \\
\hline
\end{tabular}%
}
\caption{The results of six datasets in single and mixture settings based on LLaMA2-7B, LLaMA2-13B and LLaMA-33B.}
\label{Table: scaling}
\vspace{-0.1cm}
\end{table*}

\noindent\textbf{Expert Number} \quad Considering the redundancy among experts, following \cite{gao2024higher}, we set different numbers of experts at different layers in Figure \ref{fig:expert}. Keeping the total number of experts constant, we choose three settings, i.e., (2,4,6,8), (5,5,5,5), (8,6,4,2). It is observed that assigning more experts at higher layers and fewer experts at lower layers, i.e., (2,4,6,8), works better. This is consistent with people's intuition: the lower layers of the model mainly extract general knowledge, which can be well learned by a small number of experts. While the higher layers of the model focus more on acquiring specific features of different tasks, and a larger number of experts can better capture multi-aspect capabilities.

\begin{table*}[htbp]
\centering
\resizebox{0.8\textwidth}{!}{%
\begin{tabular}{l|c|c|c||c}
\hline
\textbf{Model} & \textbf{LoRA number} & \textbf{Forward param} & \textbf{Trainable param} & \textbf{Avg\_score} \\
\hline
\textbf{FP-tuning} & / & 6.738B & 6.738B & 70.38 \\
\hline
\textbf{LoRA} & (1A+1B)*32 & 6.743B & 0.419B & 69.57 \\
\textbf{MoLA} & (5A+5B)*32 & 6.761B & 2.228B & 70.00 \\
\textbf{MoSLD} & (1A+5B)*32 & 6.572B & 1.389B & 71.56\\
\hline
\end{tabular}%
}
\caption{The number of LoRA matrices, forward parameters, and trainable parameters for FP-tuning, LoRA, MoLA, and our MoSLD during training. Here, "A" is matrix A, "B" is matrix B, and "5" is the average number of experts per layer. We also report the average results across 6 datasets under the mixture setting.}
\label{Table:Computation Efficiency}
\vspace{-0.6cm}
\end{table*}

\subsection{Mix with Other Data}
\label{sec:mixture}

\noindent\textbf{Mathematical Reasoning Data} \quad We construct a new multi-task setting, including commonsense QA task (OBQA), reading comprehension task (MCTest), subject knowledge QA task (Arc-c), and mathematical reasoning task (GSM8K). As shown in Table \ref{Table: little_commonality}, we found that for plain LoRA, the mixture setting was 1.11 points lower than the single setting on average, especially for GSM8K, it is reduced by 5.61\%, which shows that it is very challenging for plain LoRA to train multiple tasks with little commonality. However, for our MoSLD, the mixture setting is 1.18 points higher than the single setting on average. For the GSM8K with the largest difference, it is also improved by 0.67\%. This shows that MoSLD is also effective for tasks with little commonality. This is because for tasks with little commonality, although the role of the shared general-feature matrix becomes smaller, the unique-feature matrix still captures the knowledge of each task, which further shows that our MoSLD can effectively alleviate the data conflict problem in multi-task learning. 

\noindent\textbf{Mix with General Data} \quad In Figure \ref{Figure: data_amount}, we illustrate the impact of adding varying amounts of randomly filtered data from OpenOrca\footnote{https://huggingface.co/datasets/Open-Orca/OpenOrca} to the mixed dataset $D_{mix}$. The data amount from OpenOrca ranges from 1,375 to 22,000. We observed that for MoLA, as the amount of general data increases, performance initially improves before eventually declining. This suggests that mixing a large amount of general data can lead to data conflicts and domain knowledge forgetting. In contrast, MoSLD demonstrates an upward trend in performance with the increase in data amount for OBQA, MCTest, Arc-e, and Arc-c. However, performance on CSQA and Race experiences a decline. We attribute this to significant distribution differences between these datasets and the general data. Overall, our model consistently outperforms MoLA when mixing various amounts of generic data. This underscores our model's ability to effectively leverage general knowledge across different tasks. 


\subsection{Scaling of Model Size}
Table \ref{Table: scaling} shows the results of our model for the six datasets both in single and mixture settings as the model size scalings. We find that the performance of our model increases with the size of the model, whether in single or mixture settings, which is in line with our expectations. In addition, it is observed that the results improve by 1.36\%, 1.61\%, and 1.91\% from single to mixture for LLaMA2-7B, LLaMA2-13B, and LLaMA-33B, respectively. The experimental results show that our method has achieved good performance on models of different sizes, and has a certain scaling ability. We also give the model size scaling results of other LoRA-based baselines, which can be seen in the Appendix \ref{sec:scaling}.

\vspace{-0.3cm}
\subsection{Analysis of Computation Efficiency}
\label{sec:cost}

In Table \ref{Table:Computation Efficiency}, we further show the computational efficiency of our model. We first analyze the number of new LoRA modules inserted in ordinary LoRA, MoLA, and MoSLD. Since MoLA introduces the MoE framework, the trainable parameters become 5 times that of ordinary LoRA, and its results are improved by 0.43 points from 69.57 to 70.00. We believe that despite the introduction of a large number of trainable parameters, the change in results is not very large, which is a method of sacrificing efficiency for effect. In addition, we also found that although our method reduces 128 matrix A compared to MoLA, it is still 1.56\% higher than MoLA and 1.99\% higher than LoRA. This shows that although our MoSLD introduces multiple LoRAs through the MoE framework, the expert sharing mechanism greatly reduces the additional parameters and achieves a balance between effect and efficiency. We also compare FP-tuning. Athough our trainable parameters are 20.6\% of FP-tuning, but it still achieves a 1.18 point improvement. This also proves that our MoSLD is indeed an extremely efficient-parameter fine-tuning method.

\section{Conclusion}

In this paper, we propose MoSLD, which is a mixture-of-shared-LoRAs model with dropout strategy. Unlike traditional LoRA-MoE approaches, we design a sharing mechanism for matrix A, which aims to capture the general-feature among various tasks. A dropout strategy is also applied to the matrix A, solving the overfitting caused by parameter redundancy to a certain extent. Evaluations show that MoSLD outperforms the baseline in both single-task and multi-task scenarios. Especially in multi-task scenarios, where it can effectively alleviate knowledge conflict and forgetting problems. In general, our model is extremely parameter-efficient for fine-tuning.

\section*{Limitations}

Although MoSLD achieves significant improvements over existing baselines, there are still avenues worth exploring in future research. (1) This paper focuses on applying MoSLD on the matrix Q and V of the attention layer. We hope to extend this method to the FFN layer. (2) This paper explores the multi-task setting of directly mixing multiple datasets and compares with the performance of a single task. We plan to study the impact of multi-task data ratio on MoSLD. (3) This paper emphasizes the extraction of general and unique features by the upper and lower projection matrices in LoRA, and intends to visualize this phenomenon in the future.

\section*{Ethics Statement}
LoRA has emerged as a pivotal technique for refining extensive pre-trained models. Nevertheless, its efficacy tends to fail in multi-task learning. Conversely, the MoE architecture offers a promising remedy to this setback. However, it introduces hurdles such as the interference of data across diverse domains and the risk of forgetting knowledge from various tasks. Furthermore, MoE substantially inflates parameter counts, presenting computational challenges. In light of these considerations, we present MoSLD in this paper, a model that integrates the strengths of both approaches. MoSLD, a mixture-of-shared-LoRAs model with a dropout strategy, addresses these obstacles ingeniously. By sharing the upper projection matrix in LoRA among different experts, MoSLD fosters the acquisition of broad knowledge across tasks while allowing the lower projection matrix to concentrate on task-specific features. Additionally, the application of dropout mitigates parameter overfitting in LoRA. The experimental results prove the effectiveness of our model andevaluation framework. Besides, there is no hugebiased content in the datasets and the models. Ifthe knowledge base is further used, the biased con-tent will be brought into the generated responses,just like biased content posted by content creatorson the Web which is promoted by a search engine.To prevent the technology from being abused fordisinformation, we look forward to more research  effort being paid to fake/biased/offensive contentdetection and encourage developers to carefullychoose the proper dataset and content to build theknowledge base.

\bibliography{custom}

\appendix

\section{Baselines}
\label{sec:baseline}
In this section, we introduce the baselines in detail.

\textbf{Prefix-tuning \cite{li-liang-2021-prefix, zhao-etal-2022-domain}:} This method involves incorporating soft prompts into each attention layer of the Large Language Model (LLM). These soft prompts are a series of virtual tokens pre-appended to the text. During fine-tuning, the LLM remains frozen, and only the virtual tokens are optimized.

\textbf{LoRA \cite{hu2022lora}:} A popular parameter-efficient tuning approach widely used in LLM fine-tuning, LoRA leverages low-rank matrix decomposition of pre-trained weight matrices to significantly reduce the number of training parameters.

\textbf{MoLoRA \cite{zadouri2024pushing}:} A method which is a parameter-efficient MoE by uniquely combining MoE architecture with lightweight experts.

\textbf{SiRA \cite{zhu2023sirasparsemixturelow}:} A method leverages the Sparse Mixture of Expert (SMoE) and enforces the top k experts routing with a capacity limit restricting the maximum number of tokens each expert can process.ta

\textbf{MoLA \cite{gao2024higher}:} A LoRA variant with layer-wise expert allocation, MoLA flexibly assigns a different number of LoRA experts to each Transformer layer.

\textbf{MixLoRA \cite{li2024mixloraenhancinglargelanguage}:} It inserts multiple LoRA-based experts within the feed-forward network block of a frozen pre-trained dense model and employs a commonly used top-k router.

\section{Effect on Rank}
\label{sec:rank}
In this section, we add experiments on the effect of rank for our MoSLD, with r ranging from 2 to 32. Overall, the results of the six datasets did not fluctuate much, and the best value was obtained at 8 or 16. From the perspective of efficiency, 8 is indeed a suitable hyperparameter, which is also in line with the change law of LoRA's rank. The results are as shown in Table \ref{Table: rank}:

\begin{table*}[htbp]
\centering
\resizebox{0.5\textwidth}{!}{%
\begin{tabular}{l|ccccc}
\hline
\textbf{Dataset} & \textbf{r=2} & \textbf{r=4} & \textbf{r=8} & \textbf{r=16} & \textbf{r=32}\\
\hline
\textbf{OBQA} & 76.19 & 76.34 & 78.80 & 75.53 & 74.27 \\
\hline
\textbf{CSQA} & 74.35 & 75.16 & 76.43 & 77.39 & 76.62 \\
\hline
\textbf{Race} & 75.22 & 76.01 & 76.96 & 76.74 & 74.73 \\
\hline
\textbf{MCTest} & 52.28 & 54.17 & 54.42 & 54.16 & 53.52 \\
\hline
\textbf{Arc-e} & 75.51 & 76.98 & 76.70 & 75.88 & 75.63 \\
\hline
\textbf{Arc-c} & 63.28 & 64.06 & 66.13 & 66.10 & 65.87 \\
\hline
\end{tabular}%
}
\caption{The performance of our MoSLD as different rank values.}
\label{Table: rank}
\vspace{-0.6cm}
\end{table*}

\begin{table*}[]
\centering
\resizebox{0.8\textwidth}{!}{%
\begin{tabular}{l|c|c|cccccc|c}
\hline
\multicolumn{3}{l|}{\textbf{Model}} & \textbf{OBQA} & \textbf{CSQA} & \textbf{Race} & \textbf{MCTest} & \textbf{Arc-e} & \textbf{Arc-c} & \textbf{Avg} \\
\hline
\multirow{6}{*}{\textbf{LoRA}} & \multirow{2}{*}{\textbf{7B}} & single & 75.40 & 76.33 & 76.06 & 53.10 & 73.82 & 62.71 & 69.57\\
& & mixture & 72.80 & 76.30 & 78.23 & 55.67 & 70.87 & 61.00 & 69.15\\
\cline{2-10}
 & \multirow{2}{*}{\textbf{13B}} & single & 77.21 & 79.84 & 77.34 & 58.29 & 74.99 & 63.89 & 71.93\\
& & mixture & 77.98 & 78.32 & 77.83 & 55.74 & 74.05 & 64.11 & 71.34\\
\cline{2-10}
 & \multirow{2}{*}{\textbf{33B}} & single & 79.06 & 80.97 & 81.78 & 59.54 & 77.36 & 64.79 & 73.92\\
& & mixture & 79.05 & 80.02 & 82.95 & 58.27 & 75.33 & 64.88 & 73.42\\
\hline
\multirow{6}{*}{\textbf{MoLoRA}} & \multirow{2}{*}{\textbf{7B}} & single & 75.40 & 76.33 & 76.06 & 53.10 & 73.82 & 62.71 & 69.57\\
& & mixture & 72.80 & 76.30 & 78.23 & 55.67 & 70.87 & 61.00 & 69.15\\
\cline{2-10}
 & \multirow{2}{*}{\textbf{13B}} & single & 77.46 & 81.26 & 75.33 & 51.79 & 75.83 & 64.27 & 70.99\\
& & mixture & 77.95 & 82.44 & 80.25 & 54.73 & 74.21 & 62.65 & 72.04\\
\cline{2-10}
 & \multirow{2}{*}{\textbf{33B}} & single & 78.23 & 83.18 & 79.59 & 59.41 & 82.11 & 65.28 & 74.63\\
& & mixture & 77.54 & 81.35 & 81.78 & 61.62 & 82.07 & 64.35 & 74.79\\
\hline
\multirow{6}{*}{\textbf{SiRA}} & \multirow{2}{*}{\textbf{7B}} & single & 73.99 & 76.26 & 75.63 & 48.28 & 74.02 & 62.86 & 68.51\\
& & mixture & 74.34 & 76.22 & 75.04 & 52.33 & 74.98 & 63.16 & 69.35\\
\cline{2-10}
 & \multirow{2}{*}{\textbf{13B}} & single & 75.15 & 77.93 & 78.28 & 50.78 & 73.85 & 62.03 & 69.67 \\
& & mixture & 75.01 & 76.45 & 78.11 & 50.24 & 74.52 & 61.74 & 69.35\\
\cline{2-10}
 & \multirow{2}{*}{\textbf{33B}} & single & 78.99 & 81.34 & 80.03 & 53.59 & 75.78 & 64.55 & 72.38\\
& & mixture & 79.46 & 82.02 & 80.00 & 56.84 & 75.81 & 66.75 & 73.48\\
\hline
\multirow{6}{*}{\textbf{MoLA}} & \multirow{2}{*}{\textbf{7B}} & single & 74.60 & 77.23 & 75.29 & 44.90 & 72.73 & 60.80 & 67.59\\
& & mixture & 76.60 & 73.46 & 75.25 & 54.42 & 76.34 & 63.91 & 70.00\\
\cline{2-10}
 & \multirow{2}{*}{\textbf{13B}} & single & 76.82 & 80.55 & 76.87 & 48.35 & 74.84 & 63.66 & 70.18\\
& & mixture & 77.61 & 77.59 & 77.04 & 60.83 & 76.71 & 65.27 & 72.51\\
\cline{2-10}
 & \multirow{2}{*}{\textbf{33B}} & single & 80.36 & 82.94 & 79.06 & 50.88 & 76.00 & 67.06 & 72.72\\
& & mixture & 81.79 & 85.03 & 79.82 & 57.35 & 76.48 & 68.82 & 74.88\\
\hline
\multirow{6}{*}{\textbf{MixLoRA}} & \multirow{2}{*}{\textbf{7B}} & single & 75.60 & 74.83 & 75.47 & 50.88 & 74.51 & 60.10 & 68.57\\
& & mixture & 75.80 & 76.81 & 74.79 & 54.26 & 74.41 & 63.62 & 69.95 \\
\cline{2-10}
 & \multirow{2}{*}{\textbf{13B}} & single & 77.33 & 78.34 & 76.82 & 53.12 &  77.39 & 64.53 & 71.26\\
& & mixture & 76.98 & 78.05 & 77.31 & 56.88 & 78.00 & 66.92 & 72.36 \\
\cline{2-10}
 & \multirow{2}{*}{\textbf{33B}} & single & 80.57 & 81.04 & 78.99 & 55.62 & 81.25 & 67.45 & 74.15 \\
& & mixture & 80.03 & 82.87 & 79.45 & 58.98 & 79.73 & 70.87 & 75.32\\
\hline
\multirow{6}{*}{\textbf{MoSLD}} & \multirow{2}{*}{\textbf{7B}} & single & 78.40 & 75.84 & 76.08 & 53.06 & 76.35 & 61.49 & 70.20\\
& & mixture & 78.80 & 76.43 & 76.96 & 54.42 & 76.60 & 66.13 & 71.56\\
\cline{2-10}
 & \multirow{2}{*}{\textbf{13B}} & single & 81.40 & 77.95 & 78.01 & 57.86 & 78.93 & 65.05 & 73.20\\
& & mixture & 82.20 & 78.46 & 79.87 & 58.50 & 79.67 & 70.14 & 74.81\\
\cline{2-10}
 & \multirow{2}{*}{\textbf{33B}} & single & 83.93 & 81.94 & 83.27 & 65.99 & 85.10 & 68.52 & 78.05\\
& & mixture & 84.55 & 83.26 & 84.90 & 66.73 & 85.95 & 74.36 & 79.96\\
\hline

\end{tabular}%
}
\caption{The model scaling results about LLaMA2-7B, LLaMA2-13B, and LLaMA-33B of six datasets in single and mixture settings for LoRA, MoLoRA, SiRA, MoLA, MixLoRA, and our MoSLD.}
\label{Table: scaling_more}
\vspace{-0.3cm}
\end{table*}

\section{Scaling of Model Size}
\label{sec:scaling}
In this section,We add model scaling experiments on LoRA-based baselines, such as LoRA, MoLoRA, SiRA, and MoLA. We find that for each baseline, the results improve as the model size increases, among which our model MoSLD scales even better. The results are shown in Table \ref{Table: scaling_more} :

\end{document}